\documentclass{article} 
\usepackage{iclr2024_conference,times}

\usepackage{amsmath,amsfonts,bm}

\def\eqref#1{equation~\ref{#1}}

\def\1{\bm{1}}

\def\ra{{\textnormal{a}}}

\def\rx{{\textnormal{x}}}

\def\rva{{\mathbf{a}}}

\def\erva{{\textnormal{a}}}

\def\ervx{{\textnormal{x}}}

\def\rmA{{\mathbf{A}}}

\def\vmu{{\bm{\mu}}}
\def\vtheta{{\bm{\theta}}}
\def\va{{\bm{a}}}

\def\ve{{\bm{e}}}

\def\vx{{\bm{x}}}

\def\eva{{a}}

\def\mA{{\bm{A}}}

\def\mH{{\bm{H}}}
\def\mI{{\bm{I}}}
\def\mJ{{\bm{J}}}

\def\mX{{\bm{X}}}

\def\mSigma{{\bm{\Sigma}}}

\DeclareMathAlphabet{\mathsfit}{\encodingdefault}{\sfdefault}{m}{sl}
\SetMathAlphabet{\mathsfit}{bold}{\encodingdefault}{\sfdefault}{bx}{n}
\newcommand{\tens}[1]{\bm{\mathsfit{#1}}}
\def\tA{{\tens{A}}}

\def\tX{{\tens{X}}}

\def\gG{{\mathcal{G}}}

\def\sA{{\mathbb{A}}}
\def\sB{{\mathbb{B}}}

\def\sS{{\mathbb{S}}}

\def\emA{{A}}

\newcommand{\etens}[1]{\mathsfit{#1}}

\def\etA{{\etens{A}}}

\newcommand{\E}{\mathbb{E}}

\newcommand{\R}{\mathbb{R}}

\newcommand{\sigmoid}{\sigma}

\newcommand{\KL}{D_{\mathrm{KL}}}
\newcommand{\Var}{\mathrm{Var}}

\newcommand{\Cov}{\mathrm{Cov}}

\newcommand{\normltwo}{L^2}
\newcommand{\normlp}{L^p}

\newcommand{\parents}{Pa} 

\usepackage{hyperref}
\usepackage{url}
\usepackage{graphicx}
\usepackage{comment}
\usepackage{floatrow}
\usepackage{blindtext}
\usepackage{multirow}
\iclrfinalcopy 

\title{How to Parameterize Asymmetric Quantization Ranges for Quantization-Aware Training}

\author{Jaeseong You, Minseop Park, Kyunggeun Lee, Seokjun An, Chirag Patel, \& Markus Nagel \\
Qualcomm AI Research~\thanks{Qualcomm AI Research is an initiative of Qualcomm Technologies, Inc.} \\
  \texttt{\{jaeseong,minspark,kyunggeu,seokan,cpatel,markusn\}@qti.qualcomm.com}}
\begin{document}

\maketitle

\begin{abstract}
This paper investigates three different parameterizations of asymmetric uniform quantization for quantization-aware training: (1) scale and offset, (2) minimum and maximum, and (3) beta and gamma. We perform a comprehensive comparative analysis of these parameterizations’ influence on quantization-aware training, using both controlled experiments and real-world large language models. Our particular focus is on their changing behavior in response to critical training hyperparameters, bit width and learning rate. Based on our investigation, we propose best practices to stabilize and accelerate quantization-aware training with learnable asymmetric quantization ranges.
\end{abstract}

\section{Introduction}

In settings with limited low-resources, such as on-device applications or in developing countries, model efficiency is critical. Quantization serves as a practical and effective solution to this end~\citep{DBLP:journals/corr/abs-2307-02973}. In the field of deep learning, quantization refers to the method of mapping floating-point values (i.e., model weights or intermediate activations) to lower-bit integers. The benefits are two-fold: it reduces memory footprint and accelerates computation. The demand for quantization has increased as neural networks have grown in size to achieve state-of-the-art performance. Large language models (LLMs) have been a driving force behind this trend in recent years~\citep{NEURIPS2020_1457c0d6, DBLP:journals/corr/abs-2205-01068, JMLR:v24:22-1144, DBLP:journals/jmlr/LuccioniVL23, DBLP:journals/corr/abs-2203-15556, DBLP:journals/corr/abs-2302-13971}, and similar patterns are also evident across various domains~\citep{DBLP:journals/corr/abs-2303-08774, pmlr-v202-dehghani23a, DBLP:journals/corr/abs-2311-07919}. 

\begin{figure}[h]
\begin{center}
\includegraphics[width=0.95\textwidth,height=\textheight,keepaspectratio]{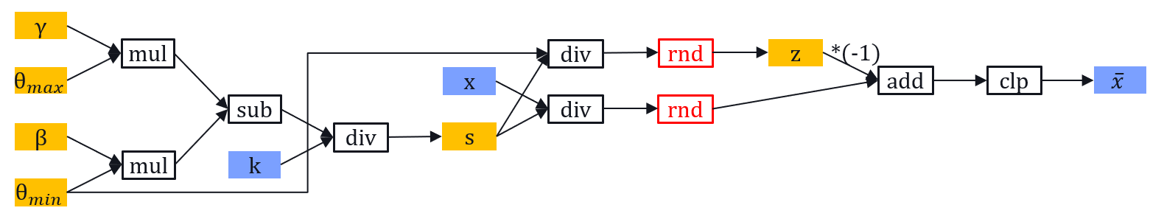}
\end{center}
\caption{Computational graph of asymmetric quantization.}
\label{fig:asymQ}
\end{figure}

Asymmetric uniform quantization and dequantization are defined as follows:
\begin{equation} \label{eq:asym_qdq}
 \begin{aligned}
    &\bar{x} = Q(x, s, z, k) = \mathrm{clip}\left(\left\lfloor\frac{x}{s}\right\rceil - \lfloor{z}\rceil, 0, k \right), \\ 
    &\hat{x} = DQ(\bar{x}, s, z) = s(\bar{x} + \lfloor{z}\rceil), \\
    &\text{where}\quad k = 2^b - 1, \quad s = \frac{\theta_{max}-\theta_{min}}{k}, \quad z = \frac{\theta_{min}}{s}.
 \end{aligned}
\end{equation}
Here, $\theta_{min}$ and $\theta_{max}$ are typically initialized to the minimum and maximum values of the input data x, and $b$ is the target bit width. While in quantization-aware-training (QAT) with learnable asymmetric quantization ranges, the standard practice is to learn $s$ and $z$~\citep{Bhalgat_2020_CVPR_Workshops}, one can opt to set other pairs of parameters as learnable, rather than $s$ and $z$ (denoted as \textit{scale/offset} hereafter). The yellow boxes in Figure~\ref{fig:asymQ} illustrate that these learnable candidates could be either $\theta_{min}$ and $\theta_{max}$ (denoted as \textit{min/max} hereafter) or $\beta$ and $\gamma$ (denoted as \textit{beta/gamma} hereafter) as well. In this paper, we (1) demonstrate that the learning patterns of these asymmetric parameterizations can be different from one another during QAT, (2) provide a comparative analysis of their differences, and (3) propose best practices to stabilize and accelerate QAT. 

\begin{figure}[h]
\begin{center}
\includegraphics[width=0.7\textwidth,keepaspectratio]{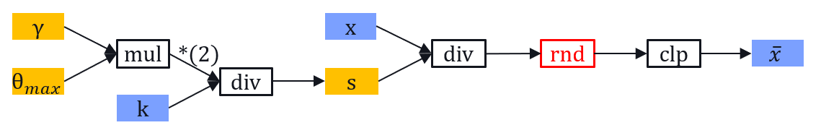}
\end{center}
\caption{Computational graph of symmetric quantization.}
\label{fig:symQ}
\end{figure}

The exploration of differences between these parameterizations is timely, as an increasing number of studies are focusing on learning-based optimization of quantization ranges, especially for extreme low-bit quantization of LLMs. These efforts include not only conventional QAT approaches~\citep{DBLP:journals/corr/abs-2305-17888, DBLP:journals/corr/abs-2308-06744, DBLP:conf/icml/Wu0AYH23}, but also quasi-QAT methods based on local---block-wise or layer-wise---optimization~\citep{DBLP:journals/corr/abs-2306-00978, DBLP:journals/corr/abs-2308-13137, DBLP:journals/corr/abs-2312-07950}. Our study has the potential to offer insights and benefits in both contexts. 

\section{Related Works}

There are two uniform quantization schema that are widely employed: symmetric (depicted in Figure~\ref{fig:symQ}) and asymmetric (depicted in Figure~\ref{fig:asymQ}). The quantization method can also be largely categorized into two: Post-training quantization (PTQ) and QAT. PTQ obtains effective quantization ranges with no (or minimal) modification of model weights~\citep{DBLP:conf/asplos/GongSZLLJMFS18, NEURIPS2019_c0a62e13}. On the other hand, QAT learns model weights with the quantization effect taken into account. This is commonly achieved by using the straight-through estimator for the non-differentiable rounding operation~\citep{DBLP:journals/corr/BengioLC13}.

In QAT with range-leaning, the quantization ranges themselves are learned, either independently with the model weights frozen \citep{DBLP:journals/corr/abs-2305-14152} or jointly with the model weights~\citep{DBLP:conf/iclr/EsserMBAM20}. The idea of learning quantization ranges was initially introduced in symmetric form~\citep{DBLP:journals/corr/abs-1805-06085, DBLP:conf/iclr/EsserMBAM20}. The concept of learnable range was then extended to asymmetric quantization. \citet{Bhalgat_2020_CVPR_Workshops} utilized the \textit{scale/offset}, while \citet{DBLP:journals/corr/abs-2201-08442} adopted the \textit{min/max}. Furthermore, \citet{DBLP:journals/corr/abs-2308-13137} introduced a novel \textit{beta/gamma} parameterization derived from \textit{min/max}. For a more in-depth understanding of quantization fundamentals, please refer to \citet{DBLP:journals/corr/abs-1806-08342, DBLP:journals/corr/abs-2106-08295, DBLP:journals/corr/abs-2201-08442}.

\section{Comparative Analysis of Asymmetric QAT Parameterizations}

In symmetric quantization, all the learnable range parameters, $s$, $\theta_{max}$ and $\gamma$, depend linearly on one another, as shown in Figure~\ref{fig:symQ}. This results in gradients that are identical except for scaling factors (see Table~\ref{tab:sym-grad} in the Appendix). However, in asymmetric quantization, the two range parameters are mutually dependent as in Figure~\ref{fig:asymQ}, resulting in complex gradients as in Table~\ref{tab:asym-grad}. See A. 2 in the Appendix why they can lead to different solutions after training.

\begin{table}[h]
\caption{Gradients of asymmetric quantization ranges.}
\label{tab:asym-grad}
\begin{center}
\begin{tabular}{c|c|c|c}
\hline
& $n < x < p$ & $x < n$ & $x > p$ \\
\hline
\hline
$\frac{d\hat{x}}{ds}$ & $\lfloor\frac{x}{s}\rceil - \frac{x}{s}$ & $n$ & $p$ \\
\hline
$\frac{d\hat{x}}{dz}$ &0 & 1 & 1 \\
\hline
\hline
$\frac{d\hat{x}}{d\theta_{min}}$ & $-\frac{1}{k}(\lfloor\frac{x}{s}\rceil - \frac{x}{s})$ & $-\frac{n}{k} + \frac{\lfloor{z}\rceil-z}{k} + 1$ & $\frac{\lfloor{z}\rceil-z}{k}$\\
\hline
$\frac{d\hat{x}}{d\theta_{max}}$ & $\frac{1}{k}(\lfloor\frac{x}{s}\rceil - \frac{x}{s})$ & $\frac{n}{k} - \frac{\lfloor{z}\rceil-z}{k}$ & $-\frac{\lfloor{z}\rceil-z}{k} + 1$ \\
\hline
\hline
$\frac{d\hat{x}}{d\beta}$ & $-\theta_{min}\frac{1}{k}(\lfloor\frac{x}{s}\rceil - \frac{x}{s})$ & $\theta_{min}(-\frac{n}{k} + \frac{\lfloor{z}\rceil-z}{k} + 1)$ & $\theta_{min}\frac{\lfloor{z}\rceil-z}{k}$\\
\hline
$\frac{d\hat{x}}{d\gamma}$  & $\theta_{max}\frac{1}{k}(\lfloor\frac{x}{s}\rceil - \frac{x}{s})$ & $\theta_{max}(\frac{n}{k} - \frac{\lfloor{z}\rceil-z}{k})$ & $\theta_{max}(-\frac{\lfloor{z}\rceil-z}{k} + 1)$ \\
\hline
\end{tabular}
\end{center}
\end{table}

\textbf{\textit{scale/offset} vs. \textit{min/max}}. Given the different QAT behaviors exhibited by the three parameterizations, the question naturally arises: which one should we use? Let us first compare \textit{scale/offset} and \textit{min/max}. One potential problem with \textit{scale/offset} is that $s$ and $z$ reside in different spaces, forming an inverse relation to one another as in \eqref{eq:asym_qdq}. Assigning identical learning rates to them would thus not be sensible, and it is unclear how to appropriately assign different rates (see the Appendix for three possible options). Another issue arises becasuse the gradients of $s$ and $z$ do not incorporate $k$, which means they cannot properly respond to changes in bit width. On the other hand, the gradients of $\theta_{min}$ and $\theta_{max}$ incorporate $k$ as in Table~\ref{tab:asym-grad}, reducing bit-width sensitivity.

An additional interesting observation about \textit{scale/offset} is that it is prone to error in situations where one of $\theta_{min}$ and $\theta_{max}$ is on its optimal point and the other is not. Once one quantization encoding reaches a local minimum, oscillation starts due to the push-and-pull between the clipping error and the quantization error. This could cause unwanted irregularities on the other encoding that has not yet converged. A good example is ReLU. While \textit{min/max} can simply fixate $\theta_{min}$ at 0 and learn only $\theta_{max}$, \textit{scale/offset} is required to move both $s$ and $z$ simultaneously at all time, which makes it more vulnerable to unstable oscillation (see Figure~\ref{fig:sz_vs_mm_relu} in the Appendix).

\begin{figure}[h]
\begin{center}
\includegraphics[width=0.48\textwidth]{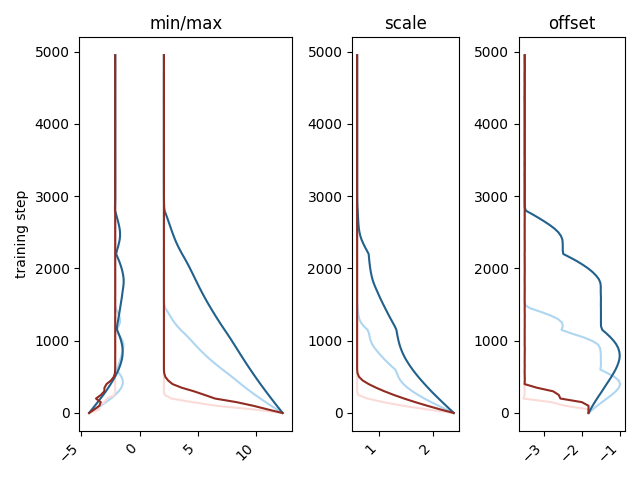}
\includegraphics[width=0.48\textwidth]{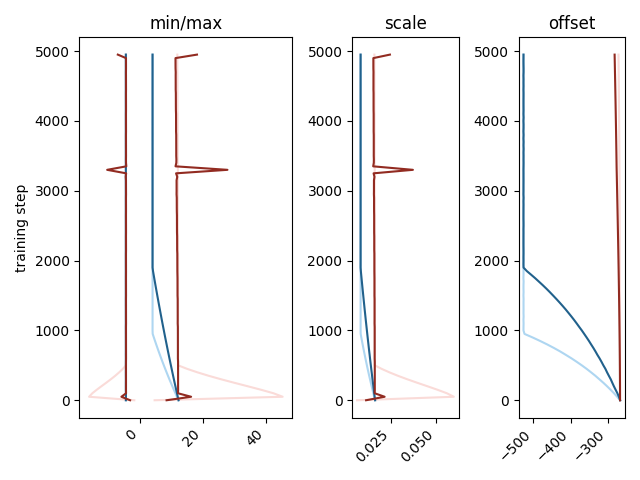}
\end{center}
\caption{Learnable ranges of \textit{scale/offset} and \textit{min/max} (x-axis) changing over 5k steps of QAT (y-axis). \textit{scale/offset} and \textit{min/max} are respectively color-coded as red and blue, and lighter shades correspond to a learning rate of 1e-2 (darker shades to that of 1e-3). The left subfigure represents 3-bit quantization (10-bit on the right). Although we experimented with 16 bit as well, \textit{scale/offset} resulted in excessively large values that could not be effectively visualized.}
\label{fig:sz_vs_mm_toy}
\end{figure}

To confirm whether the aforementioned issues indeed impede the QAT performance of \textit{scale/offset}, we perform a controlled toy experiment. We quantize a tensor of 10,000 values that follow a normal distribution. To examine bit-width sensitivity, we try low bit (3 bit) and high bit (10 bit). We also compare learning rates of 1e-2 and 5e-3, thereby ablating the impact of learning rate. The quantization range is learned to minimize the mean-squared-error (MSE) between the original tensor and the quantized-dequantized tensor:
\begin{equation} \label{eq:toy}
\underset{enc_{a},enc_{b}}{\arg\min}\frac{1}{N}\sum_{i}^{N}(DQ(Q(x_{i}, enc_{a}, enc_{b}, k), enc_{a}, enc_{b}) - x_{i})^2.
\end{equation}
Here, $enc_{a}$ and $enc_{b}$ are the learned quantization encodings (i.e. $s$ and $z$ or $\theta_{min}$ and $\theta_{max}$). We use the Adam optimizer with no weight decay~\citep{DBLP:journals/corr/KingmaB14}. The initial encoding $\theta_{min}^0$ is set to the minimum value of the tensor while $\theta_{max}^0$ is set to three times larger than the maximum value of the tensor. This is done to make the task sufficiently challenging by giving the quantizer longer asymmetric distances to manage. As observed in Figure~\ref{fig:sz_vs_mm_toy}, \textit{scale/offset} responds sensitively to the learning rate and fails to converge in the high-bit case. On the other hand, \textit{min/max} converges consistently in all scenarios.

\begin{figure}[h]
\begin{center}
\includegraphics[width=0.48\textwidth]{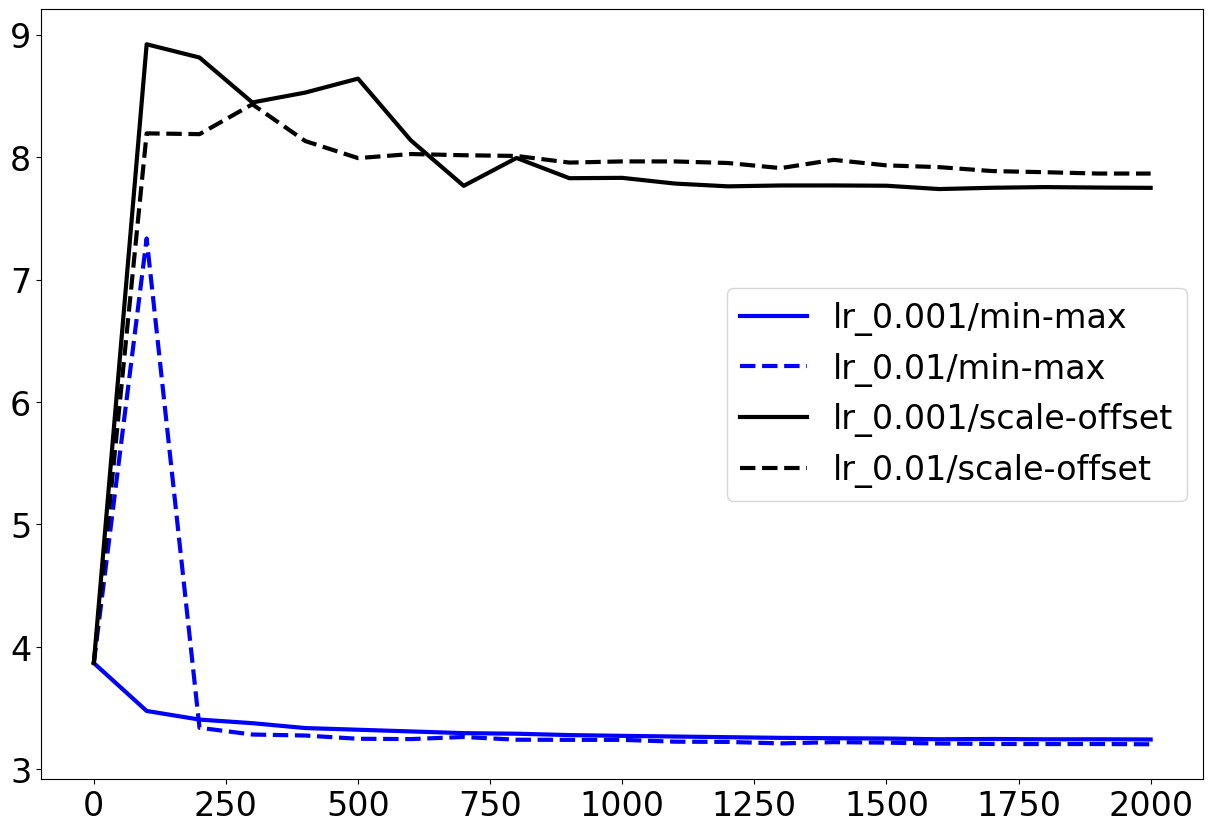}
\includegraphics[width=0.48\textwidth]{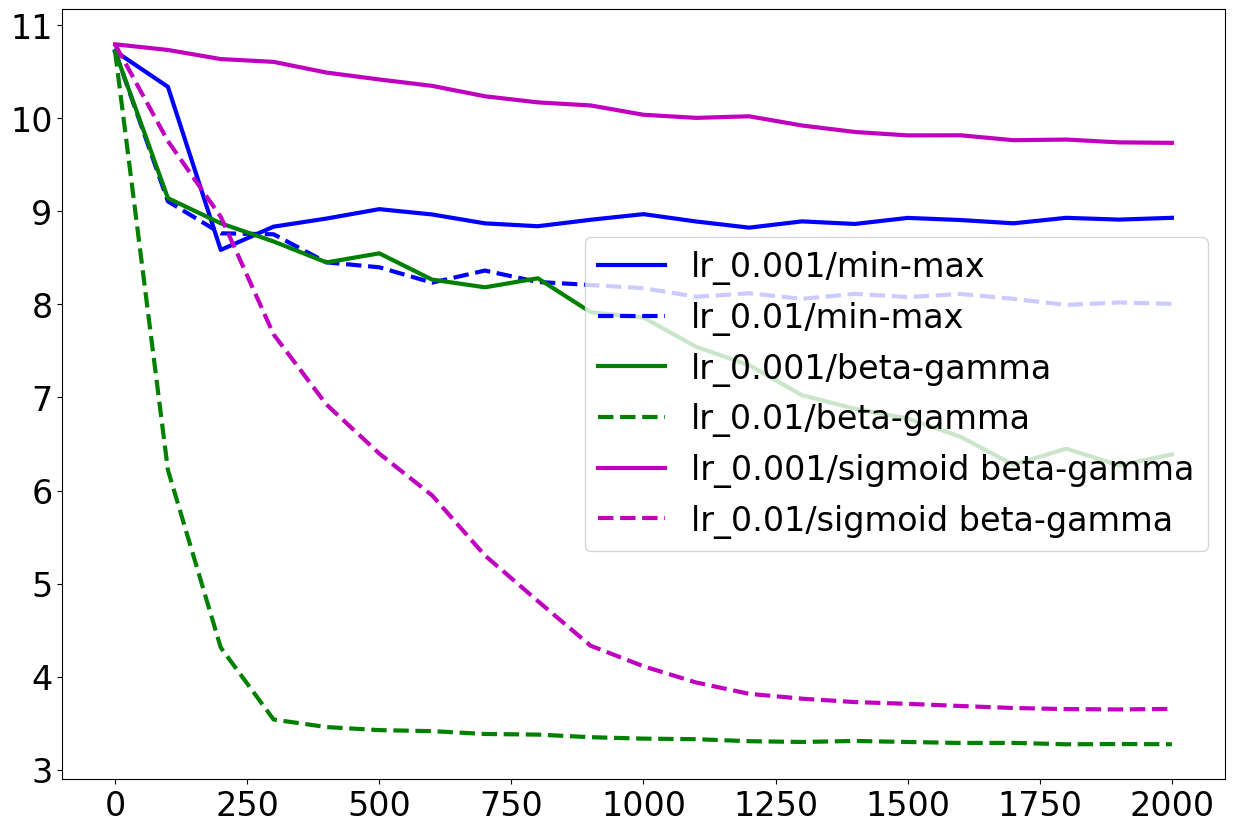}
\end{center}
\caption{Cross-entropy loss of GPT2-small QAT (y-axis) over 2k training steps (x-axis). Left depicts QAT based on \textit{min/max} and \textit{scale/offset}. Right depicts QAT based on \textit{min/max} and \textit{beta/gamma} (with and without sigmoid).}
\label{fig:sz_vs_mm_llm}
\end{figure}

Extending the comparison between \textit{scale/offset} and \textit{min/max} to a real-life scenario, we perform QAT of GPT2-small on WikiText-2~\citep{DBLP:conf/iclr/MerityX0S17}, as shown in Figure~\ref{fig:sz_vs_mm_llm}. All the weights are quantized to symmetric 4-bit integers, and all the activations are quantized to asymmetric 12-bit integers. The only exception is the layernorm weights, which follow the quantization scheme of the activations. The quantization ranges for both the weights and the activations are learned using a batch size of 8, while the model weights remain frozen. This experiment reaffirms the instability of the \textit{scale/offset} method. In contrast, \textit{min/max} reduces the cross-entropy loss consistently, irrespective of the different learning rates. We repeat the same experiment across GPT2 and OPT of different sizes as in Table~\ref{tab:big}, observing similar patterns. 

Given the apparent flaws of \textit{scale/offset}, one might find it puzzling how it has become the de-facto standard for QAT parameterization. Firstly, many QAT studies employ a symmetric quantization scheme~\citep{DBLP:conf/iclr/EsserMBAM20, DBLP:journals/corr/abs-1805-06085, DBLP:journals/corr/abs-2310-03270, DBLP:journals/corr/abs-2312-07950}, which is free from the instability of asymmetric \textit{scale/offset}. Secondly, in LLM quantization, it is often the case that only weights are quantized~\citep{DBLP:journals/corr/abs-2210-17323, DBLP:conf/iclr/FrantarAHA23, DBLP:journals/corr/abs-2308-13137, DBLP:journals/corr/abs-2312-07950}. For weight quantization, granularity is usually per-channel (as opposed to per-tensor activation quantization), and distributions tend to be symmetric with much regularized ranges compared to those of activations. Under such conditions, we find that QAT converges well regardless of parameterizations (see Figure~\ref{fig:normal} in the Appendix).

\textbf{\textit{min/max} vs. \textit{beta/gamma}}. Given its greater robustness to different bit widths/learning rates and its independent control over each of the quantization encodings, is \textit{min/max} the preferred parameterization? However, one caveat with \textit{min/max} is its slow convergence when quantization ranges must traverse large distances to reach their minima. This limitation has critical implications in practice, as studies have observed that some activations of LLM contain extremely large values~\citep{DBLP:conf/icml/XiaoLSWDH23, DBLP:journals/corr/abs-2310-08041}.

\begin{figure}[h]
\begin{center}
\includegraphics[width=0.48\textwidth]{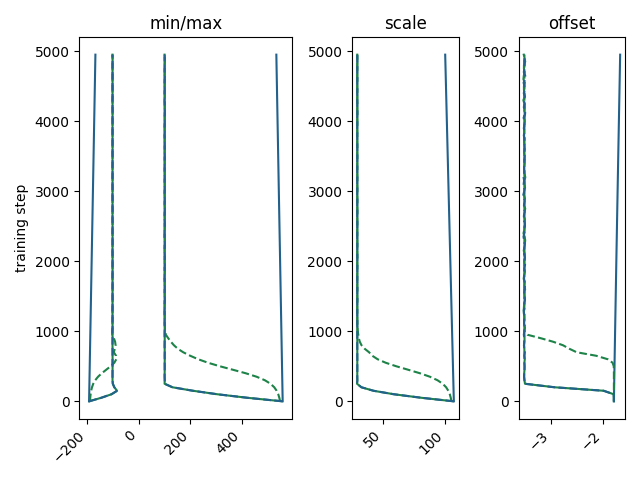}
\includegraphics[width=0.48\textwidth]{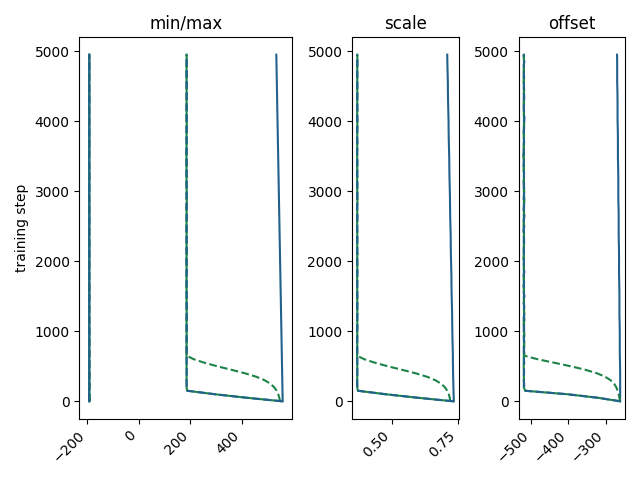}
\end{center}
\caption{Learnable ranges of \textit{min/max} and \textit{beta/gamma} changing over the course of QAT. \textit{beta/gamma} is color-coded in green (\textit{min/max} in blue). \textit{min/max+} and sigmoid-applied \textit{beta/gamma} are depicted with dashed lines. The other details of the experiment are identical to those in Figure~\ref{fig:sz_vs_mm_toy} except that we have omitted the case of $lr=1e-2$ for visual clarity.}
\label{fig:mm_vs_bg_expanded}
\end{figure}

\textit{beta/gamma} effectively overcomes this difficulty. The idea is simple. Instead of learning $\theta_{min}$ and $\theta_{max}$ themselves, new parameters $\beta$ and $\gamma$ are introduced to scale $\theta_{min}$ and $\theta_{max}$:
\begin{equation} \label{eq:beta/gamma}
 \begin{aligned}
    &s = \frac{{\gamma}\theta_{max} - {\beta}\theta_{min}}{k} \;\text{or}\; \frac{{\sigmoid(\gamma)}\theta_{max} - {\sigmoid(\beta)}\theta_{min}}{k},
    &z = \frac{{\beta}\theta_{min}}{s} \;\text{or}\; \frac{{\sigmoid(\beta)}\theta_{min}}{s}.
 \end{aligned}
\end{equation}
In Figure~\ref{fig:mm_vs_bg_expanded}, we quantize a normal distribution with a standard deviation of 50 using both \textit{min/max} and \textit{beta/gamma}. It is evident that \textit{beta/gamma} converges quickly, in stark contrast to \textit{min/max}. This is because \textit{beta/gamma} utilizes $|\theta_{min}|$ and $|\theta_{max}|$ (i.e. to scale the gradients of $\beta$ and $\gamma$, as shown in Table~\ref{tab:asym-grad}. In other words, it scales the gradients of the quantization ranges proportionally to the expected distances they need to travel (i.e. by $|\min(x_t)|$ and $|\max(x_t)|$).

As an astute reader may have noticed, \textit{beta/gamma} is highly similar to \textit{min/max} whose learning rates are scaled by $|\theta_{min}^0|$ and $|\theta_{max}^0|$ (denoted as \textit{min/max+} hereafter). Their similarity is experimentally verified in Figure~\ref{fig:mm_vs_bg_expanded}; notice how the blue dashed line (\textit{min/max+}) overlaps perfectly with the green solid line (\textit{beta/gamma} without sigmoid). They are, however, not equivalent in all cases. (1) \textit{beta/gamma} can dynamically set $\theta_{min}$ and $\theta_{max}$ to the true minimum/maximum values of $x$. Such dynamism cannot be readily attained in \textit{min/max+}. (2) \textit{beta/gamma} enables per-channel scaling of gradients by having $\theta_{min}$ and $\theta_{max}$ in vector forms. On the other hand, \textit{min/max+} requires those values to be passed as scalars to the optimizer outside the model. (3) Finally, \citet{DBLP:journals/corr/abs-2308-13137} apply a sigmoid function to $\beta$ and $\gamma$ as in \eqref{eq:beta/gamma}. Such additional treatment on the quantization encodings further differentiates \textit{beta/gamma} from \textit{min/max+}.

Let us examine these three differences. The per-channel granularity from having $\beta$, $\gamma$, $\theta_{min}$, and $\theta_{max}$ as model states is a clear advantage. The benefits of dynamic $\theta_{min}$ and $\theta_{max}$ are also evident, following the same logic as in dynamic versus static quantization. The sigmoid function on $\beta$ and $\gamma$ is, however, a double-edged sword. It stabilizes the training process, but at the cost of constraining the quantization range not to expand beyond its initial value and of slowing down the training process by compressing $\beta$ and $\gamma$. We test the impact of the sigmoid function in \textit{beta/gamma} with the controlled toy example (Figure~\ref{fig:mm_vs_bg_expanded}) and the LLM QAT (the right subfigure of Figure~\ref{fig:sz_vs_mm_llm}). In both cases, the sigmoid-free approach converges more quickly, and in the LLM experiment, it finds a lower minimum.

\begin{table}[h]
\caption{Perplexity results of LLM QAT with learned asymmetric ranges, organized by model, learning rate, and parameterization. The context length is 1024, with the exception of GPT2-XL, for which a context length of 768 is used. \textit{beta/gamma} is sigmoid-free.}
\label{tab:big}
\begin{center}
\begin{tabular}{c|c|cc|cc|cc}
\hline
 & \multirow{2}{*}{FP} & \multicolumn{2}{|c|}{\textit{scale/offset}} & \multicolumn{2}{c|}{\textit{min/max}} & \multicolumn{2}{c}{\textit{beta/gamma}} \\
& & 1e-2 & 1e-3 & 1e-2 & 1e-3 & 1e-2 & 1e-3 \\
\hline
GPT2-small & 30.0 & 2349.1   & 50256.8  & 28.6     & 28.5     & \textbf{25.6} & 27.0          \\
GPT2-XL    & 18.4 & 1712.5   & 266.6    & 17.5     & 16.6     & \textbf{15.5} & 15.8          \\
OPT-125M   & 31.8 & 4825.7   & 746.1    & 2036.1   & 52.5     & \textbf{30.6} & 31.6          \\
OPT-1.3B   & 16.8 & 3192.3   & 16.9     & 17.1     & 16.6     & 16.5          & \textbf{14.9} \\
\hline
\end{tabular}
\end{center}
\end{table}

\section{Conclusion}
Range-learning QAT is inherently unstable as it governs the rounding up/down of numerous elements by modifying a pair of quantization encodings. Adding to the complexity is our limited understanding of the impact of various parameterizations. In our efforts to stabilize and accelerate this challenging QAT process, we have made the following contributions: (1) We experimentally demonstrated that different asymmetric quantization parametrizations can behave differently during QAT. (2) We conducted a comparative analysis between \textit{scale/offset} and \textit{min/max}, demonstrating the favorable properties of the latter in terms of bit-width/learning-rate sensitivity and independent control of two quantization encodings. (3) We conducted a comparative analysis between \textit{min/max} and \textit{beta/gamma}, proposing their respective best QAT practices: \textit{min/max} with adjusted learning rates and sigmoid-free \textit{beta/gamma}.

\bibliography{iclr2024_conference}

\begin{thebibliography}{33}
\providecommand{\natexlab}[1]{#1}
\providecommand{\url}[1]{\texttt{#1}}
\expandafter\ifx\csname urlstyle\endcsname\relax
  \providecommand{\doi}[1]{doi: #1}\else
  \providecommand{\doi}{doi: \begingroup \urlstyle{rm}\Url}\fi

\bibitem[Banner et~al.(2019)Banner, Nahshan, and Soudry]{NEURIPS2019_c0a62e13}
Ron Banner, Yury Nahshan, and Daniel Soudry.
\newblock Post training 4-bit quantization of convolutional networks for rapid-deployment.
\newblock In H.~Wallach, H.~Larochelle, A.~Beygelzimer, F.~d\textquotesingle Alch\'{e}-Buc, E.~Fox, and R.~Garnett (eds.), \emph{Advances in Neural Information Processing Systems}, volume~32. Curran Associates, Inc., 2019.
\newblock URL \url{https://proceedings.neurips.cc/paper_files/paper/2019/file/c0a62e133894cdce435bcb4a5df1db2d-Paper.pdf}.

\bibitem[Bengio et~al.(2013)Bengio, L{\'{e}}onard, and Courville]{DBLP:journals/corr/BengioLC13}
Yoshua Bengio, Nicholas L{\'{e}}onard, and Aaron~C. Courville.
\newblock Estimating or propagating gradients through stochastic neurons for conditional computation.
\newblock \emph{CoRR}, abs/1308.3432, 2013.
\newblock URL \url{http://arxiv.org/abs/1308.3432}.

\bibitem[Bhalgat et~al.(2020)Bhalgat, Lee, Nagel, Blankevoort, and Kwak]{Bhalgat_2020_CVPR_Workshops}
Yash Bhalgat, Jinwon Lee, Markus Nagel, Tijmen Blankevoort, and Nojun Kwak.
\newblock Lsq+: Improving low-bit quantization through learnable offsets and better initialization.
\newblock In \emph{Proceedings of the IEEE/CVF Conference on Computer Vision and Pattern Recognition (CVPR) Workshops}, June 2020.

\bibitem[Brown et~al.(2020)Brown, Mann, Ryder, Subbiah, Kaplan, Dhariwal, Neelakantan, Shyam, Sastry, Askell, Agarwal, Herbert-Voss, Krueger, Henighan, Child, Ramesh, Ziegler, Wu, Winter, Hesse, Chen, Sigler, Litwin, Gray, Chess, Clark, Berner, McCandlish, Radford, Sutskever, and Amodei]{NEURIPS2020_1457c0d6}
Tom Brown, Benjamin Mann, Nick Ryder, Melanie Subbiah, Jared~D Kaplan, Prafulla Dhariwal, Arvind Neelakantan, Pranav Shyam, Girish Sastry, Amanda Askell, Sandhini Agarwal, Ariel Herbert-Voss, Gretchen Krueger, Tom Henighan, Rewon Child, Aditya Ramesh, Daniel Ziegler, Jeffrey Wu, Clemens Winter, Chris Hesse, Mark Chen, Eric Sigler, Mateusz Litwin, Scott Gray, Benjamin Chess, Jack Clark, Christopher Berner, Sam McCandlish, Alec Radford, Ilya Sutskever, and Dario Amodei.
\newblock Language models are few-shot learners.
\newblock In H.~Larochelle, M.~Ranzato, R.~Hadsell, M.F. Balcan, and H.~Lin (eds.), \emph{Advances in Neural Information Processing Systems}, volume~33, pp.\  1877--1901. Curran Associates, Inc., 2020.
\newblock URL \url{https://proceedings.neurips.cc/paper_files/paper/2020/file/1457c0d6bfcb4967418bfb8ac142f64a-Paper.pdf}.

\bibitem[Choi et~al.(2018)Choi, Wang, Venkataramani, Chuang, Srinivasan, and Gopalakrishnan]{DBLP:journals/corr/abs-1805-06085}
Jungwook Choi, Zhuo Wang, Swagath Venkataramani, Pierce~I{-}Jen Chuang, Vijayalakshmi Srinivasan, and Kailash Gopalakrishnan.
\newblock {PACT:} parameterized clipping activation for quantized neural networks.
\newblock \emph{CoRR}, abs/1805.06085, 2018.
\newblock URL \url{http://arxiv.org/abs/1805.06085}.

\bibitem[Chowdhery et~al.(2023)Chowdhery, Narang, Devlin, Bosma, Mishra, Roberts, Barham, Chung, Sutton, Gehrmann, Schuh, Shi, Tsvyashchenko, Maynez, Rao, Barnes, Tay, Shazeer, Prabhakaran, Reif, Du, Hutchinson, Pope, Bradbury, Austin, Isard, Gur-Ari, Yin, Duke, Levskaya, Ghemawat, Dev, Michalewski, Garcia, Misra, Robinson, Fedus, Zhou, Ippolito, Luan, Lim, Zoph, Spiridonov, Sepassi, Dohan, Agrawal, Omernick, Dai, Pillai, Pellat, Lewkowycz, Moreira, Child, Polozov, Lee, Zhou, Wang, Saeta, Diaz, Firat, Catasta, Wei, Meier-Hellstern, Eck, Dean, Petrov, and Fiedel]{JMLR:v24:22-1144}
Aakanksha Chowdhery, Sharan Narang, Jacob Devlin, Maarten Bosma, Gaurav Mishra, Adam Roberts, Paul Barham, Hyung~Won Chung, Charles Sutton, Sebastian Gehrmann, Parker Schuh, Kensen Shi, Sasha Tsvyashchenko, Joshua Maynez, Abhishek Rao, Parker Barnes, Yi~Tay, Noam Shazeer, Vinodkumar Prabhakaran, Emily Reif, Nan Du, Ben Hutchinson, Reiner Pope, James Bradbury, Jacob Austin, Michael Isard, Guy Gur-Ari, Pengcheng Yin, Toju Duke, Anselm Levskaya, Sanjay Ghemawat, Sunipa Dev, Henryk Michalewski, Xavier Garcia, Vedant Misra, Kevin Robinson, Liam Fedus, Denny Zhou, Daphne Ippolito, David Luan, Hyeontaek Lim, Barret Zoph, Alexander Spiridonov, Ryan Sepassi, David Dohan, Shivani Agrawal, Mark Omernick, Andrew~M. Dai, Thanumalayan~Sankaranarayana Pillai, Marie Pellat, Aitor Lewkowycz, Erica Moreira, Rewon Child, Oleksandr Polozov, Katherine Lee, Zongwei Zhou, Xuezhi Wang, Brennan Saeta, Mark Diaz, Orhan Firat, Michele Catasta, Jason Wei, Kathy Meier-Hellstern, Douglas Eck, Jeff Dean, Slav Petrov, and Noah Fiedel.
\newblock Palm: Scaling language modeling with pathways.
\newblock \emph{Journal of Machine Learning Research}, 24\penalty0 (240):\penalty0 1--113, 2023.
\newblock URL \url{http://jmlr.org/papers/v24/22-1144.html}.

\bibitem[Chu et~al.(2023)Chu, Xu, Zhou, Yang, Zhang, Yan, Zhou, and Zhou]{DBLP:journals/corr/abs-2311-07919}
Yunfei Chu, Jin Xu, Xiaohuan Zhou, Qian Yang, Shiliang Zhang, Zhijie Yan, Chang Zhou, and Jingren Zhou.
\newblock Qwen-audio: Advancing universal audio understanding via unified large-scale audio-language models.
\newblock \emph{CoRR}, abs/2311.07919, 2023.
\newblock \doi{10.48550/ARXIV.2311.07919}.
\newblock URL \url{https://doi.org/10.48550/arXiv.2311.07919}.

\bibitem[Dehghani et~al.(2023)Dehghani, Djolonga, Mustafa, Padlewski, Heek, Gilmer, Steiner, Caron, Geirhos, Alabdulmohsin, Jenatton, Beyer, Tschannen, Arnab, Wang, Riquelme~Ruiz, Minderer, Puigcerver, Evci, Kumar, Steenkiste, Elsayed, Mahendran, Yu, Oliver, Huot, Bastings, Collier, Gritsenko, Birodkar, Vasconcelos, Tay, Mensink, Kolesnikov, Pavetic, Tran, Kipf, Lucic, Zhai, Keysers, Harmsen, and Houlsby]{pmlr-v202-dehghani23a}
Mostafa Dehghani, Josip Djolonga, Basil Mustafa, Piotr Padlewski, Jonathan Heek, Justin Gilmer, Andreas~Peter Steiner, Mathilde Caron, Robert Geirhos, Ibrahim Alabdulmohsin, Rodolphe Jenatton, Lucas Beyer, Michael Tschannen, Anurag Arnab, Xiao Wang, Carlos Riquelme~Ruiz, Matthias Minderer, Joan Puigcerver, Utku Evci, Manoj Kumar, Sjoerd~Van Steenkiste, Gamaleldin~Fathy Elsayed, Aravindh Mahendran, Fisher Yu, Avital Oliver, Fantine Huot, Jasmijn Bastings, Mark Collier, Alexey~A. Gritsenko, Vighnesh Birodkar, Cristina~Nader Vasconcelos, Yi~Tay, Thomas Mensink, Alexander Kolesnikov, Filip Pavetic, Dustin Tran, Thomas Kipf, Mario Lucic, Xiaohua Zhai, Daniel Keysers, Jeremiah~J. Harmsen, and Neil Houlsby.
\newblock Scaling vision transformers to 22 billion parameters.
\newblock In Andreas Krause, Emma Brunskill, Kyunghyun Cho, Barbara Engelhardt, Sivan Sabato, and Jonathan Scarlett (eds.), \emph{Proceedings of the 40th International Conference on Machine Learning}, volume 202 of \emph{Proceedings of Machine Learning Research}, pp.\  7480--7512. PMLR, 23--29 Jul 2023.
\newblock URL \url{https://proceedings.mlr.press/v202/dehghani23a.html}.

\bibitem[Ding et~al.(2023)Ding, Liu, Zhang, Tu, Li, Hu, Chen, Tang, Xiong, Yin, and Wang]{DBLP:journals/corr/abs-2312-07950}
Xin Ding, Xiaoyu Liu, Yun Zhang, Zhijun Tu, Wei Li, Jie Hu, Hanting Chen, Yehui Tang, Zhiwei Xiong, Baoqun Yin, and Yunhe Wang.
\newblock {CBQ:} cross-block quantization for large language models.
\newblock \emph{CoRR}, abs/2312.07950, 2023.
\newblock \doi{10.48550/ARXIV.2312.07950}.
\newblock URL \url{https://doi.org/10.48550/arXiv.2312.07950}.

\bibitem[Esser et~al.(2020)Esser, McKinstry, Bablani, Appuswamy, and Modha]{DBLP:conf/iclr/EsserMBAM20}
Steven~K. Esser, Jeffrey~L. McKinstry, Deepika Bablani, Rathinakumar Appuswamy, and Dharmendra~S. Modha.
\newblock Learned step size quantization.
\newblock In \emph{8th International Conference on Learning Representations, {ICLR} 2020, Addis Ababa, Ethiopia, April 26-30, 2020}. OpenReview.net, 2020.
\newblock URL \url{https://openreview.net/forum?id=rkgO66VKDS}.

\bibitem[Frantar et~al.(2022)Frantar, Ashkboos, Hoefler, and Alistarh]{DBLP:journals/corr/abs-2210-17323}
Elias Frantar, Saleh Ashkboos, Torsten Hoefler, and Dan Alistarh.
\newblock {GPTQ:} accurate post-training quantization for generative pre-trained transformers.
\newblock \emph{CoRR}, abs/2210.17323, 2022.
\newblock \doi{10.48550/ARXIV.2210.17323}.
\newblock URL \url{https://doi.org/10.48550/arXiv.2210.17323}.

\bibitem[Frantar et~al.(2023)Frantar, Ashkboos, Hoefler, and Alistarh]{DBLP:conf/iclr/FrantarAHA23}
Elias Frantar, Saleh Ashkboos, Torsten Hoefler, and Dan Alistarh.
\newblock {OPTQ:} accurate quantization for generative pre-trained transformers.
\newblock In \emph{The Eleventh International Conference on Learning Representations, {ICLR} 2023, Kigali, Rwanda, May 1-5, 2023}. OpenReview.net, 2023.
\newblock URL \url{https://openreview.net/pdf?id=tcbBPnfwxS}.

\bibitem[Gong et~al.(2018)Gong, Shen, Zhang, Liu, Li, Jin, Maheshwari, Fomenko, and Segal]{DBLP:conf/asplos/GongSZLLJMFS18}
Jiong Gong, Haihao Shen, Guoming Zhang, Xiaoli Liu, Shane Li, Ge~Jin, Niharika Maheshwari, Evarist Fomenko, and Eden Segal.
\newblock Highly efficient 8-bit low precision inference of convolutional neural networks with intelcaffe.
\newblock In Luis Ceze, Natalie D.~Enright Jerger, Babak Falsafi, Grigori Fursin, Anton Lokhmotov, Thierry Moreau, Adrian Sampson, and Phillip Stanley{-}Marbell (eds.), \emph{Proceedings of the 1st on Reproducible Quality-Efficient Systems Tournament on Co-designing Pareto-efficient Deep Learning, ReQuEST@ASPLOS 2018, Williamsburg, VA, USA, March 24, 2018}, pp.\ ~2. {ACM}, 2018.
\newblock \doi{10.1145/3229762.3229763}.
\newblock URL \url{https://doi.org/10.1145/3229762.3229763}.

\bibitem[He et~al.(2023)He, Liu, Wu, Zhou, and Zhuang]{DBLP:journals/corr/abs-2310-03270}
Yefei He, Jing Liu, Weijia Wu, Hong Zhou, and Bohan Zhuang.
\newblock Efficientdm: Efficient quantization-aware fine-tuning of low-bit diffusion models.
\newblock \emph{CoRR}, abs/2310.03270, 2023.
\newblock \doi{10.48550/ARXIV.2310.03270}.
\newblock URL \url{https://doi.org/10.48550/arXiv.2310.03270}.

\bibitem[Hoffmann et~al.(2022)Hoffmann, Borgeaud, Mensch, Buchatskaya, Cai, Rutherford, de~Las~Casas, Hendricks, Welbl, Clark, Hennigan, Noland, Millican, van~den Driessche, Damoc, Guy, Osindero, Simonyan, Elsen, Rae, Vinyals, and Sifre]{DBLP:journals/corr/abs-2203-15556}
Jordan Hoffmann, Sebastian Borgeaud, Arthur Mensch, Elena Buchatskaya, Trevor Cai, Eliza Rutherford, Diego de~Las~Casas, Lisa~Anne Hendricks, Johannes Welbl, Aidan Clark, Tom Hennigan, Eric Noland, Katie Millican, George van~den Driessche, Bogdan Damoc, Aurelia Guy, Simon Osindero, Karen Simonyan, Erich Elsen, Jack~W. Rae, Oriol Vinyals, and Laurent Sifre.
\newblock Training compute-optimal large language models.
\newblock \emph{CoRR}, abs/2203.15556, 2022.
\newblock \doi{10.48550/ARXIV.2203.15556}.
\newblock URL \url{https://doi.org/10.48550/arXiv.2203.15556}.

\bibitem[Kim et~al.(2023{\natexlab{a}})Kim, Lee, Kim, Park, Yoo, Kwon, and Lee]{DBLP:journals/corr/abs-2305-14152}
Jeonghoon Kim, Jung~Hyun Lee, Sungdong Kim, Joonsuk Park, Kang~Min Yoo, Se~Jung Kwon, and Dongsoo Lee.
\newblock Memory-efficient fine-tuning of compressed large language models via sub-4-bit integer quantization.
\newblock \emph{CoRR}, abs/2305.14152, 2023{\natexlab{a}}.
\newblock \doi{10.48550/ARXIV.2305.14152}.
\newblock URL \url{https://doi.org/10.48550/arXiv.2305.14152}.

\bibitem[Kim et~al.(2023{\natexlab{b}})Kim, Lee, Lee, Hong, Chang, Sung, and Choi]{DBLP:journals/corr/abs-2308-06744}
Minsoo Kim, Sihwa Lee, Janghwan Lee, Sukjin Hong, Du{-}Seong Chang, Wonyong Sung, and Jungwook Choi.
\newblock Token-scaled logit distillation for ternary weight generative language models.
\newblock \emph{CoRR}, abs/2308.06744, 2023{\natexlab{b}}.
\newblock \doi{10.48550/ARXIV.2308.06744}.
\newblock URL \url{https://doi.org/10.48550/arXiv.2308.06744}.

\bibitem[Kingma \& Ba(2015)Kingma and Ba]{DBLP:journals/corr/KingmaB14}
Diederik~P. Kingma and Jimmy Ba.
\newblock Adam: {A} method for stochastic optimization.
\newblock In Yoshua Bengio and Yann LeCun (eds.), \emph{3rd International Conference on Learning Representations, {ICLR} 2015, San Diego, CA, USA, May 7-9, 2015, Conference Track Proceedings}, 2015.
\newblock URL \url{http://arxiv.org/abs/1412.6980}.

\bibitem[Krishnamoorthi(2018)]{DBLP:journals/corr/abs-1806-08342}
Raghuraman Krishnamoorthi.
\newblock Quantizing deep convolutional networks for efficient inference: {A} whitepaper.
\newblock \emph{CoRR}, abs/1806.08342, 2018.
\newblock URL \url{http://arxiv.org/abs/1806.08342}.

\bibitem[Kuzmin et~al.(2023)Kuzmin, Nagel, van Baalen, Behboodi, and Blankevoort]{DBLP:journals/corr/abs-2307-02973}
Andrey Kuzmin, Markus Nagel, Mart van Baalen, Arash Behboodi, and Tijmen Blankevoort.
\newblock Pruning vs quantization: Which is better?
\newblock \emph{CoRR}, abs/2307.02973, 2023.
\newblock \doi{10.48550/ARXIV.2307.02973}.
\newblock URL \url{https://doi.org/10.48550/arXiv.2307.02973}.

\bibitem[Lin et~al.(2023)Lin, Tang, Tang, Yang, Dang, and Han]{DBLP:journals/corr/abs-2306-00978}
Ji~Lin, Jiaming Tang, Haotian Tang, Shang Yang, Xingyu Dang, and Song Han.
\newblock {AWQ:} activation-aware weight quantization for {LLM} compression and acceleration.
\newblock \emph{CoRR}, abs/2306.00978, 2023.
\newblock \doi{10.48550/ARXIV.2306.00978}.
\newblock URL \url{https://doi.org/10.48550/arXiv.2306.00978}.

\bibitem[Liu et~al.(2023{\natexlab{a}})Liu, Gong, Wei, Dong, Cai, and Zhuang]{DBLP:journals/corr/abs-2310-08041}
Jing Liu, Ruihao Gong, Xiuying Wei, Zhiwei Dong, Jianfei Cai, and Bohan Zhuang.
\newblock {QLLM:} accurate and efficient low-bitwidth quantization for large language models.
\newblock \emph{CoRR}, abs/2310.08041, 2023{\natexlab{a}}.
\newblock \doi{10.48550/ARXIV.2310.08041}.
\newblock URL \url{https://doi.org/10.48550/arXiv.2310.08041}.

\bibitem[Liu et~al.(2023{\natexlab{b}})Liu, Oguz, Zhao, Chang, Stock, Mehdad, Shi, Krishnamoorthi, and Chandra]{DBLP:journals/corr/abs-2305-17888}
Zechun Liu, Barlas Oguz, Changsheng Zhao, Ernie Chang, Pierre Stock, Yashar Mehdad, Yangyang Shi, Raghuraman Krishnamoorthi, and Vikas Chandra.
\newblock {LLM-QAT:} data-free quantization aware training for large language models.
\newblock \emph{CoRR}, abs/2305.17888, 2023{\natexlab{b}}.
\newblock \doi{10.48550/ARXIV.2305.17888}.
\newblock URL \url{https://doi.org/10.48550/arXiv.2305.17888}.

\bibitem[Luccioni et~al.(2023)Luccioni, Viguier, and Ligozat]{DBLP:journals/jmlr/LuccioniVL23}
Alexandra~Sasha Luccioni, Sylvain Viguier, and Anne{-}Laure Ligozat.
\newblock Estimating the carbon footprint of bloom, a 176b parameter language model.
\newblock \emph{Journal of Machine Learning Research}, 24:\penalty0 253:1--253:15, 2023.
\newblock URL \url{http://jmlr.org/papers/v24/23-0069.html}.

\bibitem[Merity et~al.(2017)Merity, Xiong, Bradbury, and Socher]{DBLP:conf/iclr/MerityX0S17}
Stephen Merity, Caiming Xiong, James Bradbury, and Richard Socher.
\newblock Pointer sentinel mixture models.
\newblock In \emph{5th International Conference on Learning Representations, {ICLR} 2017, Toulon, France, April 24-26, 2017, Conference Track Proceedings}. OpenReview.net, 2017.
\newblock URL \url{https://openreview.net/forum?id=Byj72udxe}.

\bibitem[Nagel et~al.(2021)Nagel, Fournarakis, Amjad, Bondarenko, van Baalen, and Blankevoort]{DBLP:journals/corr/abs-2106-08295}
Markus Nagel, Marios Fournarakis, Rana~Ali Amjad, Yelysei Bondarenko, Mart van Baalen, and Tijmen Blankevoort.
\newblock A white paper on neural network quantization.
\newblock \emph{CoRR}, abs/2106.08295, 2021.
\newblock URL \url{https://arxiv.org/abs/2106.08295}.

\bibitem[OpenAI(2023)]{DBLP:journals/corr/abs-2303-08774}
OpenAI.
\newblock {GPT-4} technical report.
\newblock \emph{CoRR}, abs/2303.08774, 2023.
\newblock \doi{10.48550/ARXIV.2303.08774}.
\newblock URL \url{https://doi.org/10.48550/arXiv.2303.08774}.

\bibitem[Shao et~al.(2023)Shao, Chen, Zhang, Xu, Zhao, Li, Zhang, Gao, Qiao, and Luo]{DBLP:journals/corr/abs-2308-13137}
Wenqi Shao, Mengzhao Chen, Zhaoyang Zhang, Peng Xu, Lirui Zhao, Zhiqian Li, Kaipeng Zhang, Peng Gao, Yu~Qiao, and Ping Luo.
\newblock Omniquant: Omnidirectionally calibrated quantization for large language models.
\newblock \emph{CoRR}, abs/2308.13137, 2023.
\newblock \doi{10.48550/ARXIV.2308.13137}.
\newblock URL \url{https://doi.org/10.48550/arXiv.2308.13137}.

\bibitem[Siddegowda et~al.(2022)Siddegowda, Fournarakis, Nagel, Blankevoort, Patel, and Khobare]{DBLP:journals/corr/abs-2201-08442}
Sangeetha Siddegowda, Marios Fournarakis, Markus Nagel, Tijmen Blankevoort, Chirag Patel, and Abhijit Khobare.
\newblock Neural network quantization with {AI} model efficiency toolkit {(AIMET)}.
\newblock \emph{CoRR}, abs/2201.08442, 2022.
\newblock URL \url{https://arxiv.org/abs/2201.08442}.

\bibitem[Touvron et~al.(2023)Touvron, Lavril, Izacard, Martinet, Lachaux, Lacroix, Rozi{\`{e}}re, Goyal, Hambro, Azhar, Rodriguez, Joulin, Grave, and Lample]{DBLP:journals/corr/abs-2302-13971}
Hugo Touvron, Thibaut Lavril, Gautier Izacard, Xavier Martinet, Marie{-}Anne Lachaux, Timoth{\'{e}}e Lacroix, Baptiste Rozi{\`{e}}re, Naman Goyal, Eric Hambro, Faisal Azhar, Aur{\'{e}}lien Rodriguez, Armand Joulin, Edouard Grave, and Guillaume Lample.
\newblock Llama: Open and efficient foundation language models.
\newblock \emph{CoRR}, abs/2302.13971, 2023.
\newblock \doi{10.48550/ARXIV.2302.13971}.
\newblock URL \url{https://doi.org/10.48550/arXiv.2302.13971}.

\bibitem[Wu et~al.(2023)Wu, Li, Aminabadi, Yao, and He]{DBLP:conf/icml/Wu0AYH23}
Xiaoxia Wu, Cheng Li, Reza~Yazdani Aminabadi, Zhewei Yao, and Yuxiong He.
\newblock Understanding int4 quantization for language models: Latency speedup, composability, and failure cases.
\newblock In Andreas Krause, Emma Brunskill, Kyunghyun Cho, Barbara Engelhardt, Sivan Sabato, and Jonathan Scarlett (eds.), \emph{International Conference on Machine Learning, {ICML} 2023, 23-29 July 2023, Honolulu, Hawaii, {USA}}, volume 202 of \emph{Proceedings of Machine Learning Research}, pp.\  37524--37539. {PMLR}, 2023.
\newblock URL \url{https://proceedings.mlr.press/v202/wu23k.html}.

\bibitem[Xiao et~al.(2023)Xiao, Lin, Seznec, Wu, Demouth, and Han]{DBLP:conf/icml/XiaoLSWDH23}
Guangxuan Xiao, Ji~Lin, Micka{\"{e}}l Seznec, Hao Wu, Julien Demouth, and Song Han.
\newblock Smoothquant: Accurate and efficient post-training quantization for large language models.
\newblock In Andreas Krause, Emma Brunskill, Kyunghyun Cho, Barbara Engelhardt, Sivan Sabato, and Jonathan Scarlett (eds.), \emph{International Conference on Machine Learning, {ICML} 2023, 23-29 July 2023, Honolulu, Hawaii, {USA}}, volume 202 of \emph{Proceedings of Machine Learning Research}, pp.\  38087--38099. {PMLR}, 2023.
\newblock URL \url{https://proceedings.mlr.press/v202/xiao23c.html}.

\bibitem[Zhang et~al.(2022)Zhang, Roller, Goyal, Artetxe, Chen, Chen, Dewan, Diab, Li, Lin, Mihaylov, Ott, Shleifer, Shuster, Simig, Koura, Sridhar, Wang, and Zettlemoyer]{DBLP:journals/corr/abs-2205-01068}
Susan Zhang, Stephen Roller, Naman Goyal, Mikel Artetxe, Moya Chen, Shuohui Chen, Christopher Dewan, Mona~T. Diab, Xian Li, Xi~Victoria Lin, Todor Mihaylov, Myle Ott, Sam Shleifer, Kurt Shuster, Daniel Simig, Punit~Singh Koura, Anjali Sridhar, Tianlu Wang, and Luke Zettlemoyer.
\newblock {OPT:} open pre-trained transformer language models.
\newblock \emph{CoRR}, abs/2205.01068, 2022.
\newblock \doi{10.48550/ARXIV.2205.01068}.
\newblock URL \url{https://doi.org/10.48550/arXiv.2205.01068}.

\end{thebibliography}
\bibliographystyle{iclr2024_conference}

\appendix
\section{Appendix}

\subsection{Appropriate learning rates for \textit{scale/offset}}

Given that $s$ and $z$ exist in different spaces, it becomes necessary for QAT to adjust their gradients accordingly. A straightforward approach involves scaling them based on the absolute values of their corresponding parameters (denoted as \textit{naive} hereafter). For a more sophisticated method, we trace the implicit $\theta_{min}$ and $\theta_{max}$ to determine the amount of updates for $s$ and $z$. Let us first investigate the learning rate for $s$. The one-step updates of $\theta_{min}$ and $\theta_{max}$ with Adam optimizer are as follows:
\begin{equation} \label{eq:mm_update}
\begin{aligned}
    &s^{(t+1)} = s^{(t)} - \eta\frac{\E{[u_s]}}{{\sqrt{\E{[u_s^2]}}}} \\
    &u^{adam}_s = \frac{\E{[u_s]}}{{\sqrt{\E{[u_s^2]}}}} \\
    &s^{(t+1)} = s^{(t)} - \eta u^{adam}_s \\
    &\theta_{min}^{(t+1)} = \theta_{min}^{(t)} - \eta\frac{\E{[-\frac{1}{p}u_{s}]}}{{\sqrt{\E{[(-\frac{1}{p}u_{s})^2]}}}} \\
    &\theta_{min}^{(t+1)} = \theta_{min}^{(t)} + \eta u^{adam}_s \\
    &\theta_{max}^{(t+1)} = \theta_{max}^{(t)} - \eta u^{adam}_s \quad\text{(by the same logic)}.\\
\end{aligned}
\end{equation}
The update of $s$ under \textit{min/max} can then be expressed as:
\begin{equation} \label{eq:s_update}
\begin{aligned}
    s'^{(t+1)}  &= \frac{(\theta_{max}^{(t)} - \eta u^{adam}_{s}) - (\theta_{min}^{(t)}+ \eta u^{adam}_{s})}{k} \\
                &= \frac{(\theta_{max}^{(t)}-\theta_{min}^{(t)})}{k} - \eta \frac{2}{k}u^{adam}_s \\
                &= s'^{(t)} - \eta \frac{2}{k}u^{adam}_s.
\end{aligned}
\end{equation}
We can similarly derive a scaling factor in the case of Stochastic Gradient Descent (SGD) optimizer:
\begin{equation} \label{eq:s_update_sgd}
\begin{aligned}
s'^{(t+1)} 	&= \frac{(\theta_{max}^{(t)} - \eta \frac{1}{k}u_{s}^{sgd}) - (\theta_{min}^{(t)}+ \eta \frac{1}{k}u_{s}^{sgd})}{k} \\
			&= \frac{(\theta_{max}^{(t)}-\theta_{min}^{(t)})}{k} - \eta \frac{2}{k^2}u_{s}^{sgd} \\
			&= s'^{(t)} - \eta \frac{2}{k^2}u_{s}^{sgd}
\end{aligned}
\end{equation}

To summarize, in the case of scale, we can scale the update of $s$ under \textit{scale/offset} by $\frac{2}{k}$ to emulate the update of the derived $s$ under \textit{min/max} for Adam optimizer. The matter is, however, not straightforward for offset since the derivative of offset with respect to $\theta_{min}$ (and $\theta_{max}$) is once again complicatedly dependent on $\theta_{min}$ and $\theta_{max}$:
\begin{equation} \label{eq:dz_dmin}
\begin{aligned}
&z'^{(t+1)} = k\frac{\theta_{min} - u^{adam}_{min}}{(\theta_{max} - u^{adam}_{max})-(\theta_{min} - u^{adam}_{min})} \\
\end{aligned}
\end{equation}
One practical alternative is to use the relationship between scale and offset as defined in \eqref{eq:asym_qdq}, based on which one can scale gradient to offset as follows:
\begin{equation} \label{eq:6}
\frac{dL}{dz} = \frac{dL}{d\theta_{min}}\frac{1}{s}
\end{equation}
The proposed scaling hold regardless of optimizers, given the premise that the relationship between $z$ and $\theta_{min}$ in \eqref{eq:asym_qdq} should be maintained throughout QAT. We denote this particular scaling of learning rates for $s$ and $z$ as \textit{sophisticated} hereafter.

Besides \textit{naive} and \textit{sophisticated}, we can additionally devise a new parameterization that takes out the bit-width component $k$ out of the learnable parameters:
\begin{equation} \label{eq:ps&pz}
 \begin{aligned}
    &s' = \theta_{max}-\theta_{min}, \quad z' = \frac{\theta_{min}}{\theta_{max}-\theta_{min}}.
 \end{aligned}
\end{equation}
The scale and offset variables can then be trivially retrieved from $s'$ and $z'$ such that $s = \frac{1}{k}s'$ and $z = kz'$. With the $k$ taken out, $s'$ and $z'$ are now located in the same space, making any learning rate adjustment unnecessary (denoted as \textit{kscale/koffset} hereafter).

\begin{figure}[h]
\begin{center}
\includegraphics[width=0.48\textwidth]{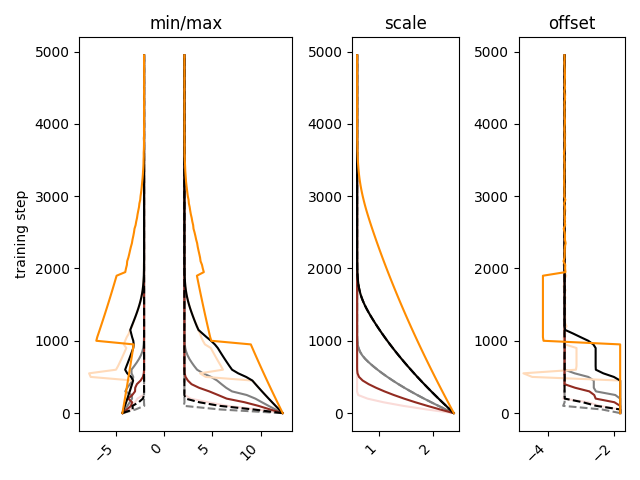}
\includegraphics[width=0.48\textwidth]{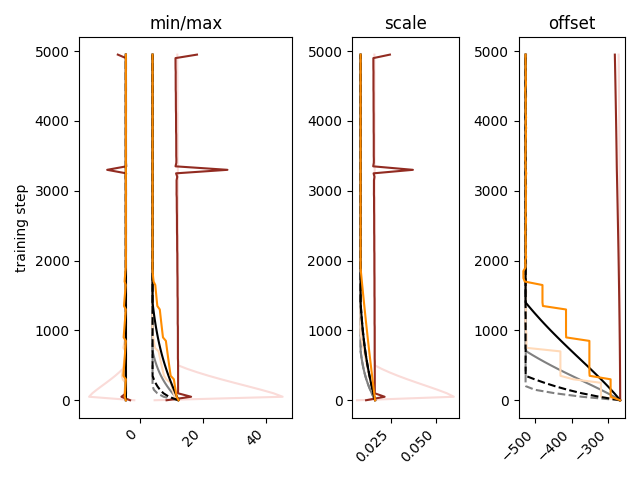}
\end{center}
\caption{Learnable ranges of (1) \textit{scale/offset} (red), (2) \textit{kscale/koffset} (orange), (3) \textit{naive} (black), and (4) \textit{sophisticated} (gray) changing over the course of QAT. The other details of the experiment are identical to those in Figure~\ref{fig:sz_vs_mm_toy}}
\label{fig:sz_asz_ksz}
\end{figure}

We perform the same experiment of quantizing a normal distribution as in Figure~\ref{fig:sz_asz_ksz} with the three aforementioned methods: \textit{naive}, \textit{sophisticated}, and \textit{kscale/koffset}. The results are illustrated in Figure~\ref{fig:sz_asz_ksz}. While all the alternatives to \textit{scale/offset} show better performance than the vanilla \textit{scale/offset} in 10-bit quantization, none shows the stability of \textit{min/max} and \textit{beta/gamma}.

\subsection{Fundamental difference between \textit{scale/offset} and \textit{min/max}}

From \eqref{eq:s_update} and \eqref{eq:dz_dmin}, we observe that the relationship between $s'^{(t)}$ and $s'^{(t+1)}$ is not the same as the relationship between $z'^{(t)}$ and $z'^{(t+1)}$. In other words, even if we make the update of the derived scale under \textit{min/max} and the update of the scale under \textit{scale/offset} identical via linear scaling, the updates of the offset will be different. After an indefinite number of updates, given $x=0$, quantization/dequantization can thus result in different answers for \textit{scale/offset} and \textit{min/max} due to their discrepancy in $z$. This is one example that evinces \textit{scale/offset} and \textit{min/max} are not one and the same. There exists no straightforward linear transformation that ensures both scale and offset undergo identical updates under \textit{min/max} and \textit{scale/offset} across their entire domain.

\subsection{Symmetric quantization}

\begin{figure}[h]
\begin{center}
\includegraphics[width=0.48\textwidth]{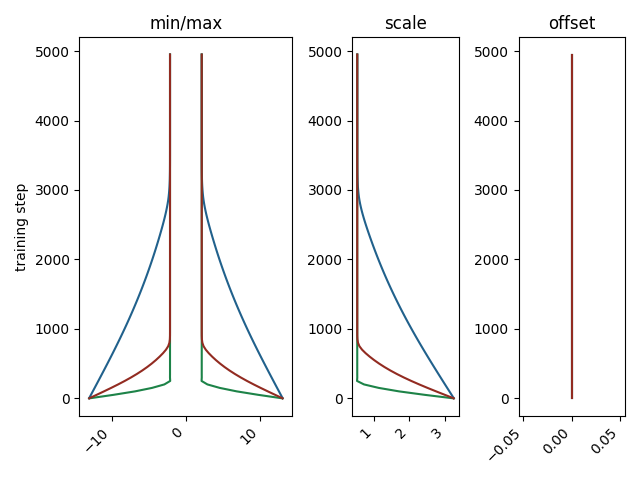}
\includegraphics[width=0.48\textwidth]{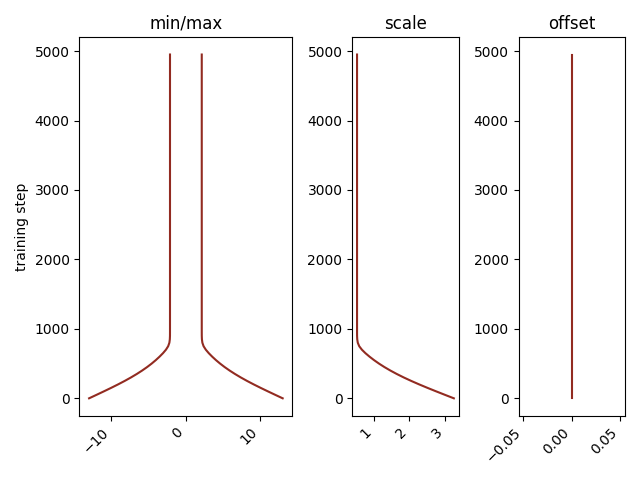}
\end{center}
\caption{Learnable ranges of \textit{scale/offset} (red), \textit{min/max} (blue), and \textit{beta/gamma} (green), changing over the course of symmetric 3-bit QAT. On the left, all parameterizations receive the same learning rate of 5e-3. On the right, the learning rates are appropriately scaled.}
\label{fig:sym}
\end{figure}

The symmetric correspondence of \eqref{eq:asym_qdq} is defined as follows:
\begin{equation} \label{eq:sym_qdq}
 \begin{aligned}
    &\bar{x} = Q(x) = \mathrm{clip}(\lfloor\frac{x}{s}\rceil, -\frac{k-1}{2}, \frac{k-1}{2}), \\ 
    &\hat{x} = DQ(\bar{x}) = s(\bar{x}), \\
    &\text{where}\quad k = 2^b - 1, \quad \theta_{max} = \max(|x|), \quad s = \frac{2 * \theta_{max}}{k}.
 \end{aligned}
\end{equation}
One of $s$, $\theta_{max}$, and $\gamma$ can be designated as a learnable parameter, as illustrated in Figure~\ref{fig:symQ}. Given the gradients of these parameters appropriately scaled as in Table~\ref{tab:sym-grad}, it is evident that they would behave identically during QAT. See Figure~\ref{fig:sym} for experimental verification, in which all the lines perfectly overlap. 

\begin{table}[h]
\caption{Gradients of symmetric quantization ranges.}
\label{tab:sym-grad}
\begin{center}
\begin{tabular}{|c|c|c|c|}
\hline
& $n < x < p$ & $x < n$ & $x > p$ \\
\hline
\hline
$\frac{d\hat{x}}{ds}$ & $\lfloor\frac{x}{s}\rceil - \frac{x}{s}$ & $n$ & $p$ \\
\hline
$\frac{d\hat{x}}{d\theta_{max}}$ & $\frac{2}{k}(\lfloor\frac{x}{s}\rceil - \frac{x}{s})$ & $\frac{2}{k}n$ & $\frac{2}{k}p$ \\
\hline
$\frac{d\hat{x}}{d\gamma}$  & $\theta_{max}\frac{2}{k}(\lfloor\frac{x}{s}\rceil - \frac{x}{s})$ & $\theta_{max}\frac{2}{k}n$ & $\theta_{max}\frac{2}{k}p$ \\
\hline
\end{tabular}
\end{center}
\end{table}

\subsection{ReLU case}

\begin{figure}[h]
\begin{center}
\includegraphics[width=0.48\textwidth]{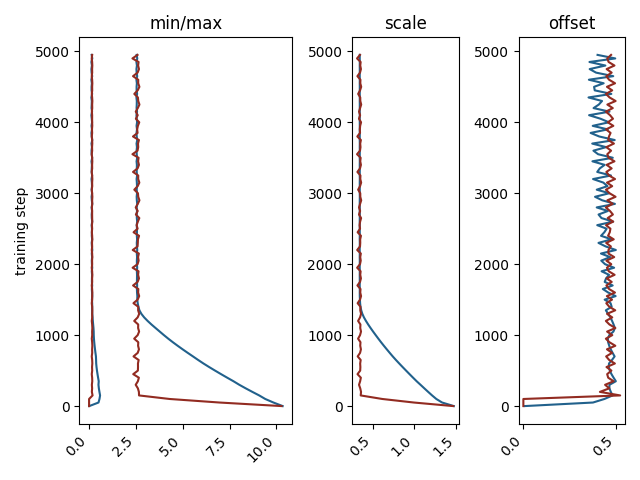}
\includegraphics[width=0.48\textwidth]{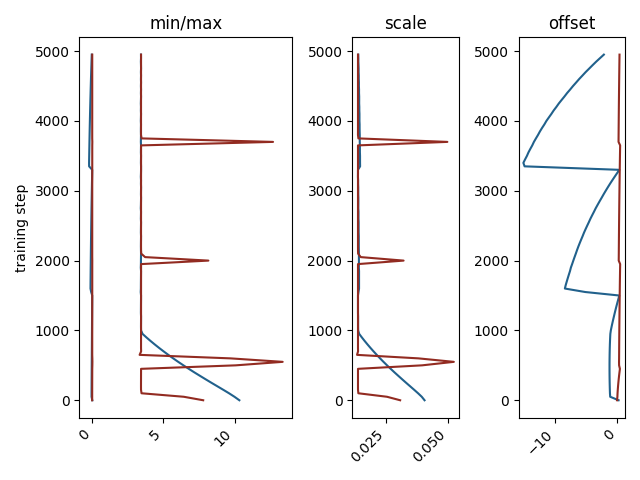}
\end{center}
\caption{Learnable ranges of \textit{scale/offset} and \textit{min/max} changing over the course of QAT. The details of the experiment are identical to those in Figure~\ref{fig:mm_vs_bg_expanded}, except that the right subfigure involves 8-bit quantization instead of 10-bit. This adjustment was made because 10-bit quantization results in values that are too large to be effectively visualized.}
\label{fig:sz_vs_mm_relu}
\end{figure}

As discussed in the main body of this work, \textit{scale/offset} is particularly unstable when one of $\theta_{min}$ and $\theta_{max}$ has already converged to its optimum and the other is still moving. This is typical of an activation after ReLU where $\theta_{min}$ is likely to be placed on the near-optimal position 0.0 from the beginning. We perform QAT on a ReLU-applied normal distribution in Figure~\ref{fig:sz_vs_mm_relu}, in which we observe severe instabilities for \textit{scale/offset}.

\subsection{Normal Quantization}

\begin{figure}[h]
\begin{center}
\includegraphics[width=0.48\textwidth]{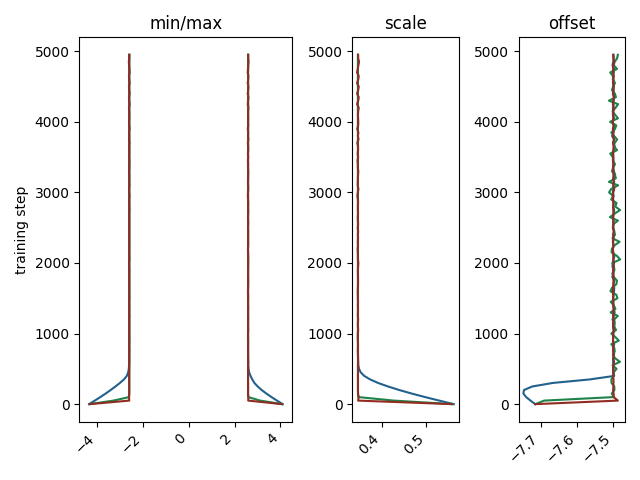}
\includegraphics[width=0.48\textwidth]{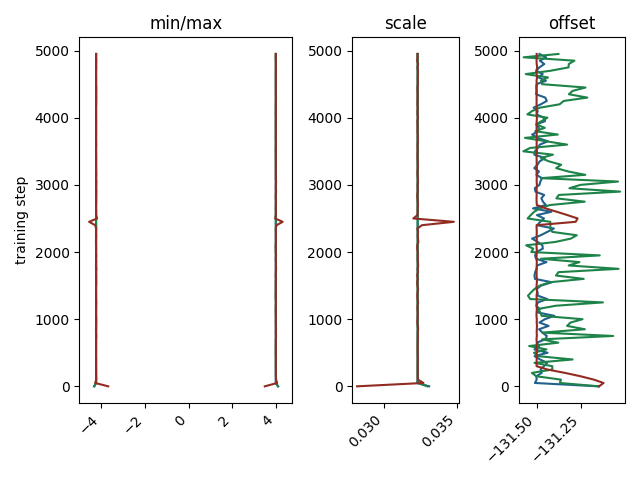}
\end{center}
\caption{Learnable ranges of \textit{scale/offset} (red) and \textit{min/max} (blue) changing over the course of QAT. The details of the experiment are identical to those in Figure~\ref{fig:sz_vs_mm_toy}, except for the initial starting points: $\theta_{min}^{0} = \min(x)$ and $\theta_{max}^{0} = \max(x)$.}
\label{fig:normal}
\end{figure}

As discussed in the main body of the paper, it might seem puzzling that there are numerous successful \textit{scale/offset} cases for QAT with learned asymmetric ranges, despite the apparent risks. To investigate whether \textit{scale/offset} can still converge successfully under less extreme conditions, we conduct an experiment, as depicted in Figure~\ref{fig:normal}, keeping the experimental setup consistent with that in Figure~\ref{fig:sz_vs_mm_toy}. However, we quantize the tensor to 4 bits and 8 bits (rather than 3 bits and 10 bits) and set $\theta_{min}$ and $\theta_{max}$ to $\min(x)$ and $\max(x)$ (instead of $\min(x)$ and $3 * \max(x)$), to alleviate the difficulty of the task. The results indicate that $\theta_{min}$ and $\theta_{max}$ of all parameterizations converge to the identical positions.

\end{document}